\documentclass[12pt]{article}
\usepackage{amsmath}
\usepackage{graphicx}
\usepackage{enumerate}
\usepackage{natbib}
\usepackage{url} 
\RequirePackage{amsmath,amsthm,amsfonts,amssymb,bm,mathrsfs}

\usepackage{dsfont,comment}
\usepackage{bm}
\usepackage{color,soul}
\usepackage{subfig}
\usepackage{hyperref}
\usepackage{multirow}
\usepackage{natbib}
\usepackage{enumitem}
\usepackage{placeins}
\usepackage[ruled,linesnumbered]{algorithm2e}
\usepackage[normalem]{ulem}

\usepackage{xr}

\newcommand{\blind}{1}

\makeatletter
\let\oldabs\abs
\def\abs{\@ifstar{\oldabs}{\oldabs*}}
\let\oldnorm\norm
\def\norm{\@ifstar{\oldnorm}{\oldnorm*}}
\makeatother

\newcommand{\ba}{\boldsymbol{a}}

\newcommand{\br}{\boldsymbol{r}}

\newcommand{\bv}{\boldsymbol{v}}

\newcommand{\bX}{\boldsymbol{X}}
\newcommand{\bx}{\boldsymbol{x}}

\newcommand{\bZ}{\boldsymbol{Z}}

\newcommand{\bbeta}{\boldsymbol{\beta}}

\newcommand{\bgamma}{\boldsymbol{\gamma}_{\cdot, j}}
\newcommand{\wbgamma}{\widehat{\boldsymbol{\gamma}}_{\cdot, j}}

\newcommand{\bSigma}{\boldsymbol{\Sigma}}

\newcommand{\bTheta}{\boldsymbol{\Theta}}

\usepackage{mathtools}
\usepackage{multirow}

\DeclareMathOperator*{\argmin}{arg\,min\;}

\theoremstyle{plain}

\newtheorem{theorem}{Theorem}[section]

\newtheorem{corollary}[theorem]{Corollary}

\theoremstyle{remark}
\newtheorem{remark}{Remark}
\newtheorem{assumption}{Assumption}

\begin{document}

\def\spacingset#1{\renewcommand{\baselinestretch}%
{#1}\small\normalsize} \spacingset{1}


\if1\blind{
	\title{\bf Adaptive Debiased Lasso in High-dimensional Generalized Linear Models with Streaming Data}
	
	\author{Ruijian $\rm{Han}^{a}$\thanks{Equally contributed authors},\hspace{0.2cm}
		Lan $\rm{Luo}^{b*}$,\hspace{0.2cm}
		Yuanhang $\rm{Luo}^{a*}$,\hspace{0.2cm}\\
		Yuanyuan $\rm{Lin}^{c}$\hspace{0.2cm}
		and
		Jian $\rm{Huang}^{a,d}$\\
		\\
		{\small {\small {$\it^{a}$  Department of Data Science and Artificial Intelligence, The Hong Kong Polytechnic University  } }}\\
		{\small {\small {$\it^{b}$ Department of Biostatistics and Epidemiology, Rutgers School of Public Health} }}\\
		{\small {\small {$\it^{c}$ Department of Statistics, The Chinese University of Hong Kong} }}\\
        {\small {\small {$\it^{d}$  Department of Applied Mathematics, The Hong Kong Polytechnic University  } }}
	}
	\maketitle
}

\fi

\if0\blind
{
  \bigskip
  \bigskip
  \bigskip
  \begin{center}
                {\Large \bf Adaptive Debiased Lasso in High-dimensional Generalized Linear Models with Streaming Data}
\end{center}
  \medskip
} \fi

\bigskip
\begin{abstract}

Online statistical inference facilitates real-time analysis of sequentially collected data, making it different from traditional methods that rely on static datasets. This paper introduces a novel approach to online inference in high-dimensional generalized linear models, where we update regression coefficient estimates and their standard errors upon each new data arrival. In contrast to existing methods that either require full dataset access or large-dimensional summary statistics storage, our method operates in a single-pass mode, significantly reducing both time and space complexity. The core of our methodological innovation lies in an adaptive stochastic gradient descent algorithm tailored for dynamic objective functions, coupled with a novel online debiasing procedure. This allows us to maintain low-dimensional summary statistics while effectively controlling the optimization error introduced by the dynamically changing loss functions. We establish the asymptotic normality of our proposed Adaptive Debiased Lasso (ADL) estimator. We conduct extensive simulation experiments to show the statistical validity and computational efficiency of our ADL estimator across various settings. Its computational efficiency is further demonstrated via a real data application to the spam email classification.

\end{abstract}

\noindent%
\vfill

\newpage
\spacingset{1.9} 
\section{Introduction}
\label{sec:intro}

Online statistical inference refers to methods and algorithms used for making inference about population parameters in a real-time environment, where data is collected sequentially rather than being static as in conventional settings. More specifically, these methods update parameter estimates and quantify their uncertainty along with sequentially accrued data information. As opposed to offline learning techniques that train over a static big dataset, online learning algorithms are expected to be computationally more efficient while achieving similar statistical performances to their offline counterparts. Most existing works in the field of online statistical inference focus on low-dimensional settings where the sample size ($n$) is much larger than the number of parameters ($p$). Their main idea is to save and update a $p$ by $p$ information matrix for quantifying the uncertainty in parameter estimation~\citep{Schifano2016CUEE,luo2020renewable,luo2022real}. Nevertheless, this idea is not directly extendable to high-dimensional settings where $p\gg n$ because the space complexity of saving such a huge matrix far exceeds that of saving the raw data. It is worth noting that most existing work in online high-dimensional settings either requires the availability of an entire raw dataset or the storage of summary statistics of large dimensions~\citep{shi2020statistical,Deshpande2019OnlineDF,EJS2182}.

In online inference, one of the main reasons to store the information matrix is that 
this quantity is pivotal in statistical inference and one can use such information directly to construct confidence interval for the target parameter(s). Moreover, online algorithms that leverage this second-order information can provide more accurate direction and step size towards the minimum, leading to faster convergence in some scenarios~\citep{Toulis2017}. However, as mentioned earlier, storing a $p\times p$ information matrix is generally not preferred in high-dimensional settings.
As  an alternative approach, the stochastic gradient descent (SGD) algorithm~\citep{robbins1951stochastic}  fits the online setting well  due to its computational and memory efficiency. In particular, SGD only requires one pass over the data, and  its space complexity can reach $\mathcal{O}(p)$. The pioneering work that uses SGD for statistical inference can be traced back to \cite{polyak1992acceleration} and this idea has also been applied to the low-dimensional models thereafter~\citep{Toulis2017, Fang2019,chen2024online}. Recently, \cite{chen2020statistical} incorporate the SGD into high-dimensional inference tasks. Specifically, they combine the regularization annealed epoch dual averaging (RADAR) algorithm, one variant of SGD \citep{agarwal2012stochastic}, with debiasing techniques \citep{Zhang_delasso_2014,vandegeer2014} in the linear models. However, their method requires access to the entire dataset in the debiasing step. Lately,~\cite{Han2023DSGD} propose a new online debiased SGD algorithm in high-dimensional linear models that is a one-pass algorithm with space complexity $\mathcal{O}(p)$. While this algorithm has many desirable properties, it is not directly applicable to high-dimensional nonlinear models such as generalized linear models. In summary, to the best of our knowledge, existing online inference methods for high-dimensional generalized linear models are either not one-pass or require a space complexity of $\mathcal{O}(p^2)$.

The goal of this paper is to develop an online statistical inference method for high-dimensional generalized linear models. In particular, the proposed method is able to sequentially update the regression coefficient estimates and their standard errors upon the arrival of a new data point, without retrieving historical raw data of space complexity $\mathcal{O}(np)$ or saving summary statistics of space complexity $\mathcal{O}(p^2)$.  It is worth mentioning that the generalized linear models are quite different from the linear models, thus the methods in \cite{chen2020statistical} and \cite{Han2023DSGD} are not easily extendable to generalized linear models. The major challenges come from two aspects. First, the objective function in the online inference framework dynamically changes as the algorithm proceeds forward. This is fundamentally different from the RADAR algorithm where the objective function is fixed. As a result, the change in the objective function introduces an additional error term which should be carefully controlled to establish asymptotic normality. Second, due to the nonlinear nature of the generalized linear models,
it is nontrivial to  develop a one-pass algorithm that utilizes summary statistics of low-space complexity. In both the first-order gradient and the Hessian matrix of the loss function, parameter estimates and data are intertwined together, making a direct decomposition infeasible.

To address these challenges, we propose a new online debiasing procedure with  two key components. First, we propose an adaptive RADAR algorithm that allows the underlying objective function to change dynamically. More importantly, the optimization error induced from the dynamically changed  objective functions is still controllable, which provides the basis to further conduct valid statistical inference. Second, to avoid reaccessing and recomputing with historical raw data, we propose to linearize the nonlinear form of the first-order gradient via Taylor approximation. Such a linearization step is highly non-trivial. It is clear that second-order approximation 
can effectively reduce approximation error, but it will involve a Hessian matrix with space complexity of $\mathcal{O}(p^2)$. Our key idea is to construct the summary statistics by combining the output from the proposed adaptive RADAR algorithm and the second-order information to reduce the dimensionality. In particular, these summary statistics require only $\mathcal{O}(p)$ space complexity but still retain the key information to conduct inference.

Our contributions can be summarized from both methodological {and theoretical aspects}. Methodologically, we propose an online inference method for high-dimensional inference, named the adaptive debiased lasso (ADL). The proposed method offers two main computational benefits. First, it is designed to operate in a single-pass mode, resulting in low time complexity. Second, this method does not require storing the raw data or large summary statistics, and its space complexity is only $\mathcal{O}(p)$, which is comparable to the complexity of storing a single data point. Thus, our method can handle ultra-high dimensional data effectively. In our real data example, we demonstrate that our method can handle datasets with millions of covariates on a standard personal computer. By contrast, alternative methods requiring access to the entire raw dataset or large summary statistics fail in similar scenarios due to memory constraints. A detailed comparison between our method and relevant existing literature is presented in Table~\ref{tab:my_label}.
{Theoretically, we establish the asymptotic normality of the proposed ADL estimator, which is an output from an online one-pass algorithm for high-dimensional generalized linear models. The asymptotic normality is built upon the consistency of two estimators, one is the lasso estimator from the RADAR algorithm and the other is the nodewise lasso estimator from the adaptive RADAR algorithm. The latter requires careful analysis of the cumulative error in a stochastic optimization procedure with a dynamically changing objective function. Moreover, unlike the conventional high-dimensional inference in offline settings, the online debiasing step involves a Taylor approximation step, and we show that the cumulative residual as a result of this approximation is negligible.}

\begin{table}[]
    \centering
        \caption{The time and space complexity of conducting  inference for a single parameter in high-dimensional models.}
    \begin{tabular}{cccc}
    \hline\hline
      Work &  Model & One-pass? & Space complexity \\
    \hline
    \cite{chen2020statistical} & LM & Yes & $\mathcal{O}(np)$  \\
    \cite{Deshpande2019OnlineDF} & LM & No & $\mathcal{O}(p^2)$   \\
    \cite{Han2023DSGD} & LM & Yes & $\mathcal{O}(p)$   \\
    \cite{shi2020statistical} & GLM & No & $\mathcal{O}(np)$  \\
    \cite{EJS2182} & GLM & No & $\mathcal{O}(p^2)$  \\
    ADL (Proposed Method) & GLM & Yes & $\mathcal{O}(p)$  \\
    \hline
       \end{tabular}\\
       Notes:  LM refers to linear model; GLM refers to generalized linear model.
        \label{tab:my_label}
\end{table}

\spacingset{1.9}

{\bf Notation}: We use $\vee$ and $\wedge$ to denote the $max$ and $min$ operator, respectively.  For a random variable $Z$,  its sub-Gaussian norm is defined as $\lVert Z \lVert_{\psi_2} = \sup_{q \geq 1} q^{-1/2} (\mathbb{E}|Z|^q)^{1/q}$; For a random
vector $\bZ$, its sub-Gaussian norm is defined as $\lVert \bZ \lVert_{\psi_2} = \sup_{\lVert \bv \lVert_2 = 1} \lVert \bv^T\bZ \lVert_{\psi_2}$. We say $\bZ$ is sub-Gaussian if its sub-Gaussian norm $\lVert \bZ \lVert_{\psi_2}$ is finite. We also call a constant $c$  the universal constant if $c$ does not depend on $n$ and $p$.  For $m\in \mathbb{N}$, we write $\{1, \cdots, m\}$ as $[m]$. 
For deterministic sequences, we use $\mathcal{O}(\cdot)$ and $o(\cdot)$ to represent the Bachmann–Landau asymptotic notation. For random sequences, we use $\mathcal{O}_{\mathbb P}(\cdot)$ and $o_{\mathbb P}(\cdot)$ in place of $\mathcal{O}(\cdot)$ and $o(\cdot)$ respectively. 
We use $\lesssim$ and $\gtrsim$ to denote the asymptotic inequality relations. For a set $S$, we use $S^c$ as its complement. For a square matrix $\bSigma$, we denote its minimum and maximum eigenvalues by $\Lambda_{\min} (\bSigma)$ and $\Lambda_{\max} (\bSigma)$ respectively.

\section{Methodology}\label{sec:method}

Suppose that $ n $ data points $ \{(\bx_i, y_i) \}_{i=1}^{n} $ arrive sequentially and each of them is an independent and identically distributed copy of $(\bX, Y)$, sampled from a generalized linear model:
\begin{equation}
\label{glm_model}
	\mathbb{P}(Y\lvert \bX; \bbeta^*) \propto \exp\left\{ \frac{Y(\bX^\intercal \bbeta^*) - \Phi(\bX^\intercal \bbeta^*)}{c(\sigma)}  \right\},
\end{equation}
where $ \Phi(\cdot) $ is a link function, {$ c(\sigma) $} is a fixed and known parameter, and $\bbeta^*$ is an $s_0$-sparse parameter, that is $\lVert \bbeta^* \lVert_0 = s_0$. Our aim is to provide a confidence interval estimator of a single coefficient or low-dimensional regression coefficients in $\bbeta^*$ along with sequentially collected data points.

In the offline setting, one of the most popular approaches is the debiased lasso method \citep{Zhang_delasso_2014,vandegeer2014}. The debiased lasso method consists of two steps: obtain the lasso estimator and correct the bias in the lasso estimator. Under the generalized linear models, we denote the negative log-likelihood function as
    $F_n(\bbeta) = ({1}/{n}) \sum_{i=1}^{n}  \left\{   -y_i(\bx^\intercal_i \bbeta) + \Phi(\bx^\intercal_i \bbeta) \right\}$,
and specify its $\ell_1$-penalized form as follows:
\begin{equation}\label{lasso}
	F_n(\bbeta; \lambda_n) = F_n(\bbeta) + \lambda_n \lVert\bbeta\lVert_1.
\end{equation} 
Then the lasso estimator is defined as 
$\widetilde{\bbeta}^{(n)} = \arg\min_{\bbeta \in \mathbb{R}^p} F_n(\bbeta; \lambda_n).
$
Since $\widetilde{\bbeta}^{(n)} $ does not have a tractable limiting distribution \citep{zhao2006model}, the debiased step is considered. Specifically, for some $j \in [p]$, the debiased estimator is calculated via 
\begin{equation}\label{offline}
  \widetilde{\bbeta}^{(n)}_{j, de}  = \widetilde{\bbeta}_{j}^{(n)} - \frac{\{ \widetilde{\boldsymbol{\gamma}}^{(n)}_{\cdot,j}\}^{\intercal} \{\nabla F_n(\widetilde{\bbeta}^{(n)})\}}{~~~\{ \widetilde{\boldsymbol{\gamma}}^{(n)}_{\cdot,j}\}^{\intercal} \{ \nabla^2 F_n(\widetilde{\bbeta}^{(n)})\}_{\cdot,j}},
\end{equation}
where $\nabla F_n(\widetilde{\bbeta}^{(n)}) $ is the gradient of the negative log-likelihood function, $\{ \nabla^2 F_n(\widetilde{\bbeta}^{(n)})\}_{\cdot,j}$ is the $j$th column of the Hessian matrix, and $\widetilde{\boldsymbol{\gamma}}^{(n)}_{\cdot,j}$ is known as the the {\em nodewise lasso estimator}. \cite{Zhang_delasso_2014} and \cite{vandegeer2014} show that such a debiased estimator converges to the normal distribution under the linear models and generalized linear models respectively. 

\begin{figure}
    \centering
    \includegraphics[width=\textwidth]{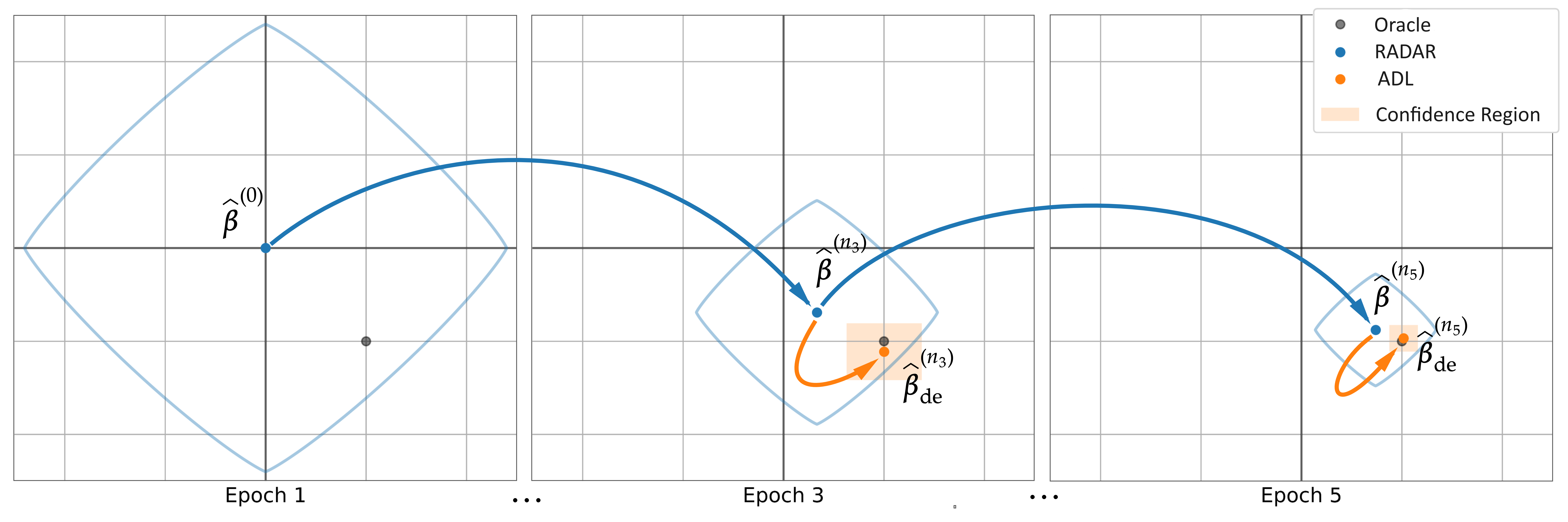}
    \caption{The whole procedure in conducting online statistical inference. }
    \label{fig:whole_trace}
\end{figure}

It is straightforward to check whether \eqref{offline} could fit into the online learning framework. Apparently, the calculations of $\widetilde{\bbeta}^{(n)}$, $\widetilde{\boldsymbol{\gamma}}^{(n)}_{\cdot,j}$,  $\nabla F_n(\widetilde{\bbeta}^{(n)}) $, and $ \{ \nabla^2 F_n(\widetilde{\bbeta}^{(n)})\}_{\cdot,j}$ all require re-accessing and recomputing with the full dataset. For this reason, we consider developing real online methods and algorithms that update the debiased lasso estimator without revisiting historical data. Based on existing literature in online lasso estimation,  an online alternative for  $\widetilde{\bbeta}^{(n)}$ is readily available \citep{agarwal2012stochastic}. Nonetheless, as indicated by~\eqref{offline}, there are {three} remaining components we need to deal with to  conduct a one-step estimation that fits into an online statistical inference framework.
In the following, we develop an online method to conduct statistical inference. We first introduce
 the RADAR algorithm used to conduct online lasso estimation in Section \ref{online_lasso}. Then,
 we develop an adaptive RADAR algorithm to find an online alternative to replace $\widetilde{\boldsymbol{\gamma}}^{(n)}_{\cdot,j}$ in  Section \ref{online_nodelasso}. Furthermore, we propose an approximated debiasing technique to replace $\nabla F_n(\widetilde{\bbeta}^{(n)}) $ and $ \{ \nabla^2 F_n(\widetilde{\bbeta}^{(n)})\}_{\cdot,j}$.  
 The resulting estimator, named adaptive debiased lasso (ADL) estimator, as well as its standard error, are presented in Sections \ref{online_debiased}--\ref{sec:variance}.
 For illustration, we present the entire procedure in Figure \ref{fig:whole_trace}. The online lasso estimation algorithm is marked in blue. Our proposed method, including the adaptive RADAR and approximated debiasing technique, is highlighted in orange.
 We wish to emphasize that our procedure is one-pass and only requires $\mathcal{O}(p)$ space complexity.

\subsection{Online lasso with RADAR}\label{online_lasso}
In this section, we introduce an online one-pass approach to finding the lasso estimator without retrieving or recomputing historical data. A natural choice that serves this purpose is the  SGD \citep{robbins1951stochastic}. Although SGD can obtain an estimator of $\bbeta^*$, its convergence rate is unsatisfactory in high-dimensional settings. 
 Specifically, \cite{shalev2011stochastic} show that the convergence rate of SGD  in high-dimensional settings with $\ell_1$-norm penalty is $\mathcal{O}(n^{-1/4})$, which is slower than $\mathcal{O}(n^{-1/2})$. Thus, interval estimation using the vanilla SGD may not have desired theoretical guarantee. In addition, ~\citet{Han2023DSGD} also show numerically that interval estimators with the vanilla SGD may not attain nominal coverage probability.
To address this issue, we adopt the regularization annealed epoch dual averaging (RADAR) algorithm \citep{agarwal2012stochastic},  a variant of the vanilla SGD.

Different from the vanilla SGD, the RADAR algorithm incorporates the multi-epoch technique. The multi-epoch technique splits a sequence of $n$ data points into $K$ epochs according to a sequence of epoch length schedule $\{T_k\}_{k\in[K]}$, where  the $k$th epoch contains data of sample size $T_k$ and $\sum_{k=1}^K T_k=n$. Let $n_k = \sum_{i=1}^{k} T_i$ denote the cumulative sample size up to the $k$th epoch. In the $k$th epoch, the RADAR algorithm runs $T_k$ rounds of the proximal SGD with dual averaging \citep{xiao2009dual} and outputs the estimate $\widehat{\bbeta}^{(n_k)} $ at the end of the epoch. The output $\widehat{\bbeta}^{(n_k)} $ is then used as an initial value in the $(k+1)$th epoch. This multi-epoch method accelerates the convergence rate by dropping the information received in the previous epochs. Since the RADAR algorithm outputs the estimate only at the end of each epoch, $\widehat{\bbeta}^{(i)}$ is well-defined only for $ i = n_k, k\in [K]$. For convenience, we define the estimates within the $(k+1)$th epoch as
\begin{equation}\label{inter_result}
    \widehat{{\bbeta}}^{(i)} = \widehat{{\bbeta}}^{(n_k)}, \ \text{if} \ n_k+1 \leq i < n_{k+1}.
\end{equation}
In other words, the estimates within each epoch remain unchanged until  the last observation in this epoch arrives. Based on \eqref{inter_result}, we obtain a sequence of the online lasso estimators, $\{\widehat{{\bbeta}}^{(i)}\}_{ i\geq n_1} $ by the RADAR algorithm. 

It is worth noting that $\widehat{{\bbeta}}^{(n)}$, the final output from the RADAR algorithm, is no longer the minimizer of $F_n(\bbeta; \lambda_n)$ defined in \eqref{lasso}. In fact, without accessing the entire dataset, it is hard to achieve the optimality as the offline estimator.
Recall $\widetilde{\bbeta}^{(n)}$ is the minimizer of $F_n(\bbeta; \lambda_n)$. Instead of the statistical error $\lVert \widetilde{\bbeta}^{(n)} - \bbeta^* \lVert_1$, we  need to consider the optimization error from the RADAR algorithm, that is $\lVert \widehat{{\bbeta}}^{(n)} - \bbeta^* \lVert_1$. Under some regularity conditions, we can conclude that the optimization error $\lVert \widehat{{\bbeta}}^{(n)} - \bbeta^* \lVert_1$ achieves the same rate as the statistical error $\lVert \widetilde{\bbeta}^{(n)} - \bbeta^* \lVert_1$ in terms of the sample size $n$.
A detailed description of the RADAR algorithm is included in the supplementary material.

\subsection{Online nodewise lasso with adaptive RADAR}\label{online_nodelasso}
Given the online lasso estimator, we still need to find an online alternative to $\widetilde{\boldsymbol{\gamma}}^{(n)}_{\cdot,j}$ for the one-step estimation \eqref{offline}. In an offline setting, the nodewise lasso estimator is computed via 
\begin{equation}\label{offline_gamma}
	 \argmin_{\br \in \mathbb{R}^{{p}}, \br_j = -1 } \frac{1}{2n} \sum_{i=1}^{n}    \left[ \ddot{\Phi}(\bx_i^\intercal \widetilde{\bbeta}^{(n)}) \{ (\bx_i)^\intercal \br\}^2 \right] + \lambda_n' \lVert\br \lVert_1.
\end{equation}
 To solve \eqref{offline_gamma} in an online fashion, one may again adopt the SGD approach, such as the RADAR algorithm.
However, according to \eqref{offline_gamma}, the stochastic gradient with respect to $\br_{-j}$  contains the term
$(\bx_i)_{-j}\{ (\bx_i)^\intercal \br \} \ddot{\Phi}(\bx_i^\intercal \widetilde{\bbeta}^{(n)}), \ i \in [n]$,
where $\widetilde{\bbeta}^{(n)}$ is not available when $i < n$. Thus, the RADAR algorithm is not directly applicable to solving the nodewise lasso problem. 

To address this issue, we modify the RADAR algorithm by replacing  $\widetilde{\bbeta}^{(n)}$ with the latest online lasso estimator of $\bbeta^*$. Since the resulting gradient is adaptive to the sequentially changed lasso estimator of $\bbeta^*$, we call this algorithm the adaptive RADAR algorithm.  Note that the initial estimator of $\bbeta^*$ is only available after processing the first $n_1$ data. Thus, we start the online nodewise lasso after the first $n_1$ data points.  In particular, we consider the epoch length $\{T_k'\}_{k \in [K']}$ and define the cumulative length or sample size as 
$ n_{k}' =  n_1 +  \sum_{j=1}^{k} T_j' $
for $k \in [K']$. Our newly proposed approach computes the stochastic gradient of $\br_{-j}$ via 
\begin{equation}\label{nodewise_lasso_SGD}
      (\bx_i)_{-j}\{ (\bx_i)^\intercal \br \} \ddot{\Phi}(\bx_i^\intercal \widehat{\bbeta}^{(n_{k-1}')}) + \lambda_k' \textit{sign}(\br_{-j}), \ \ \  n_{k-1}' < i \leq n_k',
\end{equation}
in the $k$th epoch where $\widehat{\bbeta}^{(n_{k-1}')}$ is defined in \eqref{inter_result} and \textit{sign} is the sign function. Then, we adopt \eqref{nodewise_lasso_SGD} as the stochastic gradient and follow the rest of the procedure in the RADAR algorithm. As a result, we obtain the output at the end of $k$th epoch,  denoted by ${\wbgamma}^{(n_{k}')}$ for $k \in [K']$. Similar to \eqref{inter_result}, we also define
\begin{equation}\label{inter_result_nodewise}
        {\wbgamma}^{(i)} = {\wbgamma}^{(n_k')}, \ \text{if} \ n_k' + 1 \leq i < n_{k+1}',
\end{equation}
for $k \in [K'-1]$. Then, we treat $\{ {\wbgamma}^{(i)}\}_{i \geq n_1'}$ as online nodewise lasso estimators.

The adaptive RADAR algorithm is more general than the original RADAR algorithm. Specifically, let $ \mathcal{F}_{k}' = \sigma \left( \{(\bx_i, y_i) \}_{i=1}^{n_k'} \right) $, that is the $\sigma$-field generated from $\{(\bx_i, y_i) \}_{i=1}^{n_k'}$, and
$	\bar{G}_k(\br) =    \mathbb{E} \left(  \ddot{\Phi}(\bX^\intercal  \widehat{\bbeta}^{(n_{k-1}')}  ) (\bX^\intercal \br  )^2  \big\lvert  \mathcal{F}_{k-1}'  \right)$.
In the $k$th epoch, the stochastic gradient defined in \eqref{nodewise_lasso_SGD}  comes from the objective function
\begin{equation}\label{optim}
      \bar{G}_k(\br) + \lambda_k'\lVert\br\lVert_1,
\end{equation}
which contains two parts: the loss function $\bar{G}_k(\br)$ and the regularization term $\lambda_k'\lVert\br\lVert_1$. 
We wish to highlight that  both terms vary across different epochs. In contrast, the objective function in the original RADAR algorithm \citep{agarwal2012stochastic}  is
\begin{equation}\label{optim_original}
     G(\br) + \lambda_k'\lVert\br\lVert_1,
\end{equation}
for some fixed loss function $G(\cdot)$. Compared to \eqref{optim_original},  an innovative  aspect of our proposed objective function \eqref{optim} is that,   the loss function can vary as the optimization algorithm proceeds, though it  brings new challenges in the theoretical justification.  In particular, changing the loss function also changes the target value of the optimization problem, which induces additional error.  Under certain regularity conditions, with suitable choices of hyper-parameters, we show that the overall estimation error of the resulting estimator can achieve  $\mathcal{O}_{\mathbb{P}}(n^{-1/2})$ as detailed in Theorem \ref{bgamma_oracle}.

\subsection{Approximated debiasing techniques}\label{online_debiased}
In generalized linear models, computing $ \nabla F_n(\widehat{\bbeta}^{(n)}) $ is infeasible in an online setting.
Specifically, $ \nabla F_n(\widehat{\bbeta}^{(n)})$ takes the following form
\begin{equation}\label{gradient}
	\nabla F_n(\widehat{\bbeta}^{(n)}) = \frac{1}{n} \sum_{i=1}^{n} \bx_i\left\{-y_i + \Dot{\Phi}(\bx^\intercal_i \widehat{\bbeta}^{(n)})\right\},
\end{equation}
which requires retrospective calculations with the full dataset. Due to the non-linearity of the function $ \Dot{\Phi}(\cdot)$, the computation of  $\sum_{i=1}^{n} \bx_i\dot{\Phi}(\bx^\intercal_i \widehat{\bbeta}^{(n)})$ in \eqref{gradient}  is impractical if we cannot retrieve the historical data $\{\bx_i\}_{i<n}$ after obtaining $\widehat{\bbeta}^{(n)}$. For instance, in  logistics regression, where $\Phi(t) = \{1/(1 + \exp(-t))\} $, there is no such a summary statistics to obtain $\sum_{i=1}^{n} \bx_i\dot{\Phi}(\bx^\intercal_i \widehat{\bbeta}^{(n)})$.

Instead of computing the exact value of $\sum_{i=1}^{n} \bx_i\dot{\Phi}(\bx^\intercal_i \widehat{\bbeta}^{(n)})$, we will approximate it. 
A straightforward solution is the plug-in approximation, namely
 $\sum_{i=1}^{n} \bx_i\dot{\Phi}(\bx^\intercal_i \widehat{\bbeta}^{(i)})$.   
 
 However, complications arise from the approximation error in the plug-in method, which involves an additional residual term of $\mathcal{O}_{\mathbb{P}}\left(n^{-1 / 2}\right)$. Consequently, we cannot derive asymptotic normality.

 \begin{remark}

To establish the asymptotic normality of the debiased estimator, the estimator can typically be decomposed into the average of $n$ independent and identically distributed (i.i.d.) random variables, with residual terms being $o_{\mathbb P}(n^{-1/2})$. However, when using the plug-in method in our online procedure, the residual term is $\mathcal{O}_{\mathbb P}(n^{-1/2})$ rather than  $o_{\mathbb P}(n^{-1/2})$ under our theoretical framework.  Consequently, the asymptotic normality cannot be established by the central limit theorem. We include the ablation studies in the supplementary material to provide numerical evidence to show the advantages of our method over the plug-in method.
 \end{remark}

 To obtain the residual term being ${o}_{\mathbb P}(n^{-1/2}),$  we employ the approximated debiasing techniques via Taylor approximation. Specifically,
  \begin{equation}
     \dot{\Phi}(\bx^\intercal_i \widehat{\bbeta}^{(n)}) = \dot{\Phi}(\bx^\intercal_i \widehat{\bbeta}^{(i)}) + \ddot{\Phi}(\bx^\intercal_i \widehat{\bbeta}^{(i)})\bx^\intercal_i(\widehat{\bbeta}^{(n)} - \widehat{\bbeta}^{(i)}) + \mathcal{O}(\lVert \widehat{\bbeta}^{(n)} - \widehat{\bbeta}^{(i)} \lVert_2^2),\  i \in [n].
 \end{equation}
 The key idea of our method is to use the second-order approximation rather than the plug-in approximation, namely,
 \begin{equation*}
     \sum_{i=1}^{n} \bx_i\dot{\Phi}(\bx^\intercal_i \widehat{\bbeta}^{(n)}) \approx \sum_{i=1}^{n}\left\{ \bx_i\dot{\Phi}(\bx^\intercal_i \widehat{\bbeta}^{(i)}) + \bx_i\ddot{\Phi}(\bx^\intercal_i \widehat{\bbeta}^{(i)})\bx^\intercal_i(\widehat{\bbeta}^{(n)} - \widehat{\bbeta}^{(i)})\right\}.
 \end{equation*}
Nonetheless, the Hessian matrices $\bx_i\ddot{\Phi}(\bx^\intercal_i \widehat{\bbeta}^{(i)})\bx^\intercal_i, \ i \in [n]$ are $p \times p$ matrices, which require a large amount of memory capacity. To ensure the superiority of our method in terms of space complexity,  we construct the following summary statistics rather than storing the Hessian matrices. {Letting $l = \min \{{L}: n_{L} \geq n_1' \}$, we compute}
\begin{align}
     &\ba_{1}^{(m)}  =  \sum_{i=n_l+1}^{m}  \bx_i\left\{ - y_i + \dot{\Phi}(\bx^\intercal_i \widehat{\bbeta}^{(i)}) \right\}, \ \nonumber
 \ba_2^{(m)}  = \sum_{i=n_l+1}^{m}\bx_i^\intercal {\wbgamma}^{(i)} \ddot{\Phi}(\bx^\intercal_i \widehat{\bbeta}^{(i)})\bx_i,\\
    & a_{3}^{(m)} = \sum_{i=n_l+1}^{m} \bx_i^\intercal {\wbgamma}^{(i)} \ddot{\Phi}(\bx^\intercal_i \widehat{\bbeta}^{(i)})\bx^\intercal_i\widehat{\bbeta}^{(i)},\
      {a}_4^{(m)} =\sum_{i=n_l+1}^{m}  \bx_i^\intercal {\wbgamma}^{(i)} (\bx_i)_j \left\{\ddot{\Phi}(\bx_i^\intercal \widehat{\bbeta}^{(i)} )\right\}, \label{summary_statistics}
 \end{align}
 where $\ba_1^{(m)} $ and $\ba_2^{(m)} $ are $p$-dimensional vectors while the other elements $a_{3}^{(m)}$ and $ a_{4}^{(m)}$ are scalars. The above summary statistics are computed  from $ i = n_l + 1$ rather than $i = 1$,  ensuring the existence of ${\wbgamma}^{(i)}$ and $\widehat{\bbeta}^{(i)}$. 

 Based on the statistics defined in \eqref{summary_statistics}, we approximate $ \{ \widetilde{\boldsymbol{\gamma}}^{(m)}_{\cdot,j}\}^{\intercal} \{\nabla F_m(\widetilde{\bbeta}^{(m)})\}
$ in \eqref{offline} by
 \begin{align*}
\{\ba^{(m)}_1\}^\intercal {\wbgamma}^{(m)}  +    \{\ba^{(m)}_2\}^\intercal \widehat{{\bbeta}}^{(m)}   - a_{3}^{(m)}.
\end{align*}
 Moreover, we use ${a}_4^{(m)}$, the $j$th element of ${\ba}_2^{(m)}$, as an alternative to $\{ \widetilde{\boldsymbol{\gamma}}^{(m)}_{\cdot,j}\}^{\intercal} \{ \nabla^2 F_m(\widetilde{\bbeta}^{(m)})\}_{\cdot,j}$. 

\begin{figure}
    \centering
    \includegraphics[width=\textwidth]{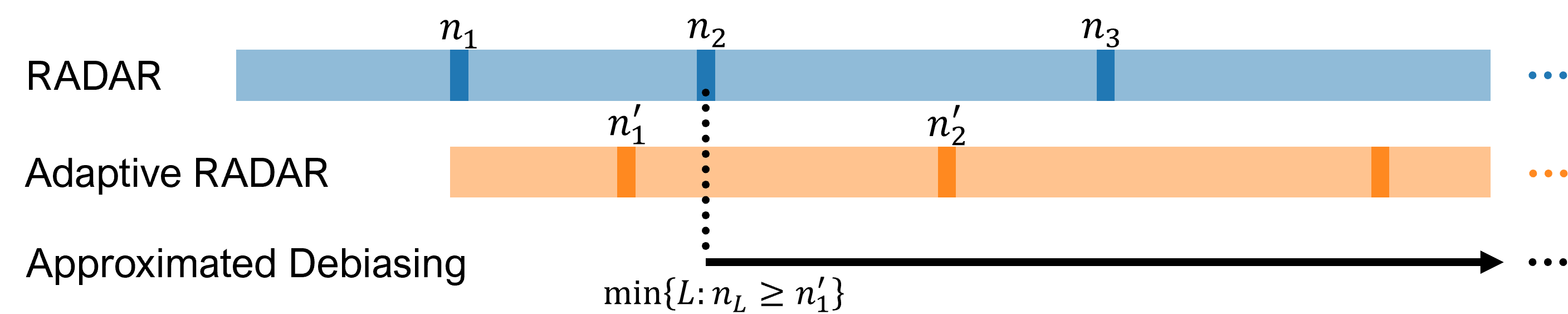}
    \caption{The timeline of three key steps in constructing the ADL estimator.}
    \label{fig:ADL}
\end{figure}

 Consequently, we define the adaptive debiased lasso (ADL) estimator as
\begin{equation}\label{approx_debiased}
	{\widehat{\bbeta}}^{(m)}_{j, \text{de}} := \widehat{{\bbeta}}^{(m)}_{j} - \frac{ \{\ba^{(m)}_1\}^\intercal {\wbgamma}^{(m)}  +    \{\ba^{(m)}_2\}^\intercal \widehat{{\bbeta}}^{(m)}   - a_{3}^{(m)}  }{{a}_4^{(m)}}, \ m > n_{l}.
\end{equation}
\noindent One key advantage of \eqref{approx_debiased} is that we can update these summary statistics recursively instead of storing the raw data in memory. For example, when the $m$th data arrives, we compute
 \begin{equation}\label{online_update}
    \ba_{1}^{(m)} =  \ba_{1}^{(m-1)} + \bx_m\left\{ - y_m + \dot{\Phi}(\bx^\intercal_m \widehat{\bbeta}^{(m)}) \right\}.
\end{equation}
Then, following the update as in \eqref{online_update}, the calculation of  ${\widehat{\bbeta}}^{(m)}_{j, \text{de}}$ in \eqref{approx_debiased} is one-pass with $\mathcal{O}(p)$ space complexity. For ease of illustration, a diagram that depicts the timeline of the three key steps in online estimation and inference procedure is presented in Figure \ref{fig:ADL}.

\subsection{Variance and confidence interval}\label{sec:variance}
To construct a confidence interval, we need to further find the variance of ${\widehat{\bbeta}}^{(m)}_{j, \text{de}}$. 
To this end,  we design another scalar-valued summary statistic
\begin{equation*}
   a_5^{(m)} = \sum_{i=n_{l}+1}^{m}  (\bx_i^\intercal {\wbgamma}^{(i)} )^2 \left\{\dot{\Phi}(\bx^\intercal_i \widehat{\bbeta}^{(i)}) - y_i\right\}^2.
\end{equation*}
Then the estimated standard error of the ADL estimator is computed as
\begin{eqnarray}\label{approx_debiased_var}
\widehat{\nu}^{(m)}_j = \sqrt{a_5^{(m)}}/ |a_4^{(m)}|. 
\end{eqnarray}
Consequently, based on \eqref{approx_debiased} and \eqref{approx_debiased_var}, the $(1-\alpha)$-confidence interval for $\bbeta_j^*$ with $0 < \alpha < 1$ at time $m$ is given by:
\begin{equation}\label{confidence_interval_2}
	\left({\widehat{\bbeta}}^{(m)}_{j, \text{de}}-  z_{\alpha/2}\widehat{\nu}^{(m)}_j,  {\widehat{\bbeta}}^{(m)}_{j, \text{de}} + z_{\alpha/2}\widehat{\nu}^{(m)}_j\right),\ m > n_l,
\end{equation}
where $z_{\alpha/2}$ is the  $\{1- (\alpha/2)\}$-quantile of the standard normal distribution. We present the pseudocode of the ADL in Algorithm \ref{debiased_algorithm_2}.  In numerical studies, we take the zero point as the initial value for both $\widehat{\bbeta}^{(0)}$ and ${\wbgamma}^{(0)}$. Algorithm \ref{debiased_algorithm_2} processes each data point only once and does not store any original raw data. In addition, the required summary statistics are at most $p$-dimensional. Therefore, Algorithm \ref{debiased_algorithm_2} is one-pass and only requests $\mathcal{O}(p)$ space complexity, as stated in Table \ref{tab:my_label}.

\spacingset{1.6}
  \begin{center}
\begin{algorithm}
\SetKwInOut{Input}{Input}
\caption{Pseudocode of the proposed ADL.}\label{debiased_algorithm_2}
\Input{Streaming data $\{(\bx_m, y_m)\}_{m=1}^{n}$,  
epoch schedule $\{T_k\}_{k \in [K]}$ and $\{T_k'\}_{k \in [K']}$, 
regularization parameters $\{\lambda_k\}_{k \in [K]}$ and $\{\lambda_k'\}_{k \in [K']}$, significance level $\alpha$, and index of the variable of interest $j$.}
Initialize $\widehat{\bbeta}^{(0)}, {\wbgamma}^{(0)}, \ba_1^{(0)}, \ba_2^{(0)}, a_3^{(0)}, a_4^{(0)}, a_5^{(0)}$\;
\For{$m = 1, \ldots, n$}{Receive the data point $(\bx_m, y_m)$\;
Conduct the RADAR with the hyper-parameters $\{T_k, \lambda_k \}_{k \in [K]}$  \;
Conduct the adaptive RADAR with the hyper-parameters $\{T_k', \lambda_k'\}_{k \in [K']}$ \;
Obtain the online estimator $\widehat{\bbeta}^{(m)}$ and ${\wbgamma}^{(m)}$ via \eqref{inter_result} and \eqref{inter_result_nodewise} respectively\;
Update $\ba_1^{(m)}, \ba_2^{(m)}, a_3^{(m)}, a_4^{(m)}, a_5^{(m)}$ recursively as~\eqref{online_update}\;
Compute $ {\widehat{\bbeta}}^{(m)}_{j, \text{de}} $ and $ \widehat{\nu}^{(m)}_j $ via \eqref{approx_debiased} and \eqref{approx_debiased_var} respectively\;
Output the $(1-\alpha)$-confidence interval \\
 \qquad  \qquad \qquad \qquad \qquad  $\left({\widehat{\bbeta}}^{(m)}_{j, \text{de}}-  z_{\alpha/2}\widehat{\nu}^{(m)}_j,  {\widehat{\bbeta}}^{(m)}_{j, \text{de}} + z_{\alpha/2}\widehat{\nu}^{(m)}_j\right)$ \;
 \vspace{0.01in}
 Update  $ \widehat{\bbeta}^{(m)}, {\wbgamma}^{(m)}, \ba_1^{(m)}, \ba_2^{(m)}, a_3^{(m)}, a_4^{(m)}, a_5^{(m)} $ and clear the raw data in memory\;}
\end{algorithm}
  \end{center}
\spacingset{1.9}
\section{Theoretical properties}\label{sec:thm}
\subsection{Oracle inequalities}
To show the consistency of the output $ \widehat{\bbeta}^{(n)} $ from the RADAR algorithm, we need the following assumptions.
\begin{assumption}\label{assump_1}
The predictor vector
$\bX$ is a zero-mean and sub-Gaussian random vector, whose sub-Gaussian norm is $\kappa_1$ and the covariance matrix $\bSigma$ satisfies: $0 < M_1^{-1} \leq \Lambda_{\min} (\bSigma) \leq  \Lambda_{\max} (\bSigma) \leq M_1 < \infty $ for a universal constant $M_1$.
\end{assumption}
\begin{assumption}\label{assump_2}
  The response $Y$ is a sub-Gaussian random variable with the sub-Gaussian norm $\kappa_2$. In addition, the second derivative of the link function is {Lipschitz} and uniformly bounded by a universal constant $M_2$, that is, $\max_{t \in \mathbb{R}}\lvert \ddot{\Phi}(t) \lvert \leq M_2$.
\end{assumption}
\begin{assumption}\label{assump_3}
	Assume  $\lVert \bbeta^* \lVert_0 = s_0$ and $\lVert \bbeta^* \lVert_2 = M_3$ for a universal constant $M_3$.
\end{assumption}

Assumption \ref{assump_1} considers the sub-Gaussian random design, which is commonly adopted in establishing the oracle inequality for the high-dimensional model in the literature. Assumption \ref{assump_2} is a mild regularity condition on the generalized linear models defined in \eqref{glm_model}. It is satisfied by a broad class of models, such as linear  model with sub-Gaussian error and logistic model. Assumption \ref{assump_3} is a basic assumption on the parameter space in the sparse regression settings.

Let $d_0 = \lVert \widehat{\bbeta}^{(0)} - \bbeta^* \lVert_1$ denote the initial error of the algorithm. Without loss of generality, we assume that $d_0\gtrsim 1.$ Then, the following theorem establishes the convergence rate  of the output from the RADAR algorithm.
\begin{theorem}\label{bbeta_oracle}
Suppose that Assumptions \ref{assump_1}--\ref{assump_3} hold. With the choice of hyper-parameters according to (S.1)--(S.3)  in the supplementary material, the following events
	\begin{equation}\label{bbeta_oracle_eqn}
		\lVert \widehat{\bbeta}^{(i)} - \bbeta^* \lVert_1 \leq  C_1s_0d_0\sqrt{\frac{(\log p)^3}{i}},\ \	\lVert \widehat{\bbeta}^{(i)} - \bbeta^* \lVert_2 \leq C_1d_0 \sqrt{ \frac{s_0(\log p)^3}{i}}
	\end{equation}
	for $ i \geq n_1 $ hold uniformly for a universal constant $C_1$ with probability at least $1 - 7(\log p)^{-6}.$
\end{theorem}

Compared to the offline approach~\citep{12STS400}, our error bound  has two additional terms:  the first is the initial error, $d_0$. This is because our online algorithm is one-pass and iterates through its input data exactly once. This means that the number of iterations equals to the 
size of the input dataset, which is finite in practice. In this regard, the initial error is not negligible and should be taken into consideration. 
Another  term is an additional  logarithmic factor $(\log p)$, which arises from the optimization process in the RADAR algorithm.   Unlike the offline approach,  we have to further handle the optimization error. This additional error can be seen as a trade-off between online and offline approaches:  we reduce time and space complexity at the cost of some additional estimation error.

Compared to the online high-dimensional approach~\citep{Han2023DSGD}, Theorem \ref{bbeta_oracle} achieves the same error rate under the sub-Guassian design. However, our work focuses on the generalized linear models, whereas \cite{Han2023DSGD} focuses only on the linear models.

\begin{remark}
	If we revise the sub-Gaussian design in Assumption \ref{assump_1} to the uniformly bounded design, that is, $\lVert \bX \lVert_\infty $ is bounded by some universal constant,  the convergence rate in terms of $\ell_1$-norm will be $\mathcal{O}_\mathbb{P} ( s_0d_0 \sqrt{(\log p)/i})$ for $i \geq n_1$. But
	 the uniformly bounded design contradicts our simulation settings where  $\bX$ follows the normal distribution. Hence,  the theoretical result under the uniformly bounded design is not presented.
\end{remark}

Next, we consider the estimation error of the output from the adaptive RADAR. Let  ${\bTheta}(\bbeta) = [\mathbb{E}\{\bX^\intercal \ddot{\Phi}(\bX^\intercal \bbeta)\bX\} ]^{-1}$ represent the inverse of the Hessian matrix, and let $\overline{\bTheta}(\bbeta) = [\text{diag}\{{\bTheta}(\bbeta) \}]^{-1} {\bTheta}(\bbeta) $ denote its normalized version. Two additional assumptions on the normalized inverse Hessian matrix are imposed.
\begin{assumption}\label{inverse_hessian_assumption_sparse}
	There exists a universal constant $M_4$ such that $0 < M_4^{-1} \leq \Lambda_{\min} (\overline{\bTheta}(\bbeta^*)) \leq  \Lambda_{\max} (\overline{\bTheta}(\bbeta^*)) \leq M_4 < \infty $.  The columns of $\overline{\bTheta}(\bbeta^*)$ are $s_1$-sparse, that is $\max_{j \in [p]} \lVert \{\overline{\bTheta}(\bbeta^*)\}_{\cdot,j} \lVert_0 $ \ $ = s_1.$ 
\end{assumption}
\begin{assumption}\label{inverse_hessian_assumption}
		There is a universal constant $M_5$ such that $\lVert \overline{\bTheta}(\bbeta) - \overline{\bTheta}(\bbeta^*) \lVert_1 \leq M_5 \lVert \bbeta - \bbeta^* \lVert_1 $, for any $\bbeta \in \{\bbeta: \lVert \bbeta - \bbeta^*  \lVert_2 \leq 2M_3, \lVert \bbeta - \bbeta^*  \lVert_1 \leq 2d_0  \}$, where $M_3 =   \lVert \bbeta^* \lVert_2.$
\end{assumption}

Assumption \ref{inverse_hessian_assumption_sparse} ensures the $\ell_0$-sparsity of $\bgamma^*$ and the boundedness of $\lVert \bgamma^* \lVert_2$ as in Assumption \ref{assump_3}. This assumption is commonly adopted in establishing the oracle inequality of nodewise lasso, such as \cite{vandegeer2014}, \cite{belloni2016post} and \cite{10.1214/16-AOS1448}. Assumption \ref{inverse_hessian_assumption} requires  the normalized inverse Hessian matrix to be a {local} Lipschitz operator, which is a new requirement compared with those in the literature of the nodewise lasso. Nonetheless, this assumption is similar to Assumption (D4) in \cite{vandegeer2014} to some extent. The former could be regarded as the population-level version of the latter.
Note that Assumption \ref{inverse_hessian_assumption} holds automatically  under the high-dimensional linear models since $\overline{\bTheta}(\bbeta)$ degenerates to a constant matrix.

Similarly, let $d_1 = \lVert {\wbgamma}^{(0)} - \bgamma^* \lVert_1$ be
 the initial error. We provide the convergence rate of the output from the adaptive RADAR algorithm in the next theorem.
\begin{theorem}\label{bgamma_oracle}
	Suppose that Assumptions \ref{assump_1}--\ref{inverse_hessian_assumption} holds. {With the choice of hyper-parameters according to (S.4)--(S.6) in the supplementary material}, given any fixed $j \in [p]$, the following events 
	\begin{equation}\label{gamma_error}
		\lVert {\wbgamma}^{(i)} - \bgamma^* \lVert_1 \leq  C_2\left\{  s_1d_1 + (s_1 + s_0)d_0 \right\}\sqrt{\frac{(\log p)^3}{i - n_1}}
	\end{equation}
 	for $ i \geq n_1' $ hold uniformly for a universal constant $C_2$ with probability at least $1 - 14(\log p)^{-6}$.
\end{theorem}
{In contrast to  Theorem \ref{bbeta_oracle}, the error bound in \eqref{gamma_error} not only involves the initial error $d_1$ and the sparsity  $s_1$ of the normalized inverse Hessian matrix, but also the error from the online lasso algorithm via $d_0$ and $s_0$. This phenomenon shows the difference between the RADAR algorithm and the adaptive RADAR algorithm. As shown in \eqref{optim}, the loss function in the adaptive RADAR algorithm dynamically changes based on the most recent estimates from the online lasso procedure. In other words, the target values are constantly changing. Therefore, an additional error will be introduced from the drifted target values.  With Assumption \ref{inverse_hessian_assumption} and the careful design of the hyper-parameters (S.4)--(S.6) in the supplementary material, we are able to control this additional error. Consequently, the overall estimation error will converge at the rate of $\mathcal{O}_{\mathbb P}(n^{-1/2})$.

\subsection{Asymptotic normality}
By  Theorems \ref{bbeta_oracle} and \ref{bgamma_oracle}, we establish the asymptotic normality of  ${\widehat{\bbeta}}^{(n)}_{j, \text{de}}$ defined in \eqref{approx_debiased}.

\begin{theorem}\label{asy_normal}
Suppose that Assumptions \ref{assump_1}--\ref{inverse_hessian_assumption} hold  and 
    \begin{align}\label{final_requirement}
    ( s_0s_1d_0d_1  +  s_0^2d_0^2 + s_0s_1d_0^2 )  (\log p)^4 \log_2(n) = o(\sqrt{n}).
    \end{align}
     With the choice of hyper-parameters according to (S.1)--(S.6) in the supplementary material, given any fixed $j \in [p],$ 
  \begin{align*}
 \frac{1}{\widehat{\nu}^{(n)}_j }({\widehat{\bbeta}}^{(n)}_{j, \text{de}} -  {{\bbeta}}^{*}_{j}) \to  \mathcal{N}(0, 1) \ \text{ in distribution} \ \ \text{ as } n \to \infty .
\end{align*}
\end{theorem}
Compared to existing work in high-dimensional inference, our theoretical results are different in two aspects. On one hand, the assumptions in Theorem \ref{asy_normal} are imposed at the population level. In contrast, most of the existing inference methods \citep{vandegeer2014,10.1214/16-AOS1448,ma2021global,shi2020statistical, EJS2182} rely on the bounded individual probability condition.
For example, in logistic regression, the bounded individual probability condition takes the form
 $\mathbb{P}(y_i=1|\bx_i) \in (c,1-c) $  for $i \in [n]$ and some positive universal constant $c$. 
{This difference in assumptions is primarily due to stochastic optimization in our work, where the objective function is defined at the population level. Consequently, we do not rely on assumptions at the sample level.
When $\bx_i, i \in [n]$ follows a normal distribution with a slightly larger variance,  the bounded individual probability condition may not be satisfied. To the best of our knowledge, except for  \cite{guo2021inference} and \cite{cai2023statistical}, most existing methods require the bounded individual probability condition. Therefore, both our method and the one proposed in \cite{cai2023statistical} can still provide valid inference under  Gaussian design, as evidenced in our numerical studies. 
}

On the other hand, we require slightly stronger conditions in terms of the sparsity levels $s_0$ and $s_1$, sample size $n$, and the dimension of features $p$. This is the trade-off between online and offline algorithms. Specifically, the online estimator may not be the optimizer  of some objective functions. Hence, some optimality conditions, such as the Karush–Kuhn–Tucker (KKT) condition, cannot be used. Therefore, we need to impose slightly stronger conditions compared to its offline counterpart so as to establish the asymptotic properties. Similar consideration can be also found in the existing  theoretical results related to online high-dimensional inference, such as Theorem 4.10 in \cite{Deshpande2019OnlineDF} and Theorem 1 in \cite{Han2023DSGD}.

We end this section by providing the validity of the confidence interval for $\bbeta^*_j$ given in Algorithm \ref{debiased_algorithm_2}
in the next corollary. 

\begin{corollary}
Under the conditions in Theorem \ref{asy_normal}, for any fixed $j \in [p]$ and $\alpha \in (0,1)$, 
\begin{align*}
\mathbb{P}\left\{\bbeta^*_j \in \big({\widehat{\bbeta}}^{(n)}_{j, \text{de}} -  z_{\alpha/2}\widehat{\nu}^{(n)}_j,  {\widehat{\bbeta}}^{(n)}_{j, \text{de}} + z_{\alpha/2}\widehat{\nu}^{(n)}_j \big)   \right\} \to 1-\alpha  \ \text{ as } n \to \infty.
\end{align*}

\end{corollary}

\section{Simulation studies}\label{sec:sim}

In this section, we conduct simulation studies to examine the finite-sample performance of the ADL in high-dimensional logistic regression. 
All experiments, including the real data example in Section \ref{sec:real}, are performed with an 11th Gen Intel Core i7-11370H processor and 16GB memory storage. 

{\bf Simulation settings:} We randomly generate a sequence of $n$ independent and identically distributed copies $\{(\bx_i, y_i) \}_{i=1}^{n}$  from the logistic regression model
$\mathbb{P}(y = 1\lvert \bx; \bbeta^*)  =  1 - \mathbb{P}(y = 0\lvert \bx; \bbeta^*) = {\exp(\bx^\intercal\bbeta^*) }/\{1+ \exp(\bx^\intercal\bbeta^*)\}$,
 where $\bx \sim \mathcal{N}(0, \bSigma)$. Recall that the cardinality of the support set of $\bbeta^{*}$ is $s_0$. We divide the support set into two halves $\{S_1, S_2\}$, with true coefficients  $\bbeta^{*}_{S_1}=1$ and $\bbeta^{*}_{S_2}=-1$ respectively. Both $S_1$ and $S_2$ are randomly selected.
 We conduct two experiments: one is the comparison between the ADL and some existing methods to show the advantage of the ADL in terms of time complexity. Another is used to indicate the low space complexity of the ADL. 
 
 In our first experiment, we choose $n=200, p=500$ and $s_0=6$. We compare the performance of the ADL with offline debiased lasso (deLasso) in \cite{vandegeer2014}, link-specific weighting (LSW) method proposed in \cite{cai2023statistical} and online debiased lasso (ODL) proposed in \cite{EJS2182}. In addition, we check the robustness of these methods with respect to violating the bounded individual probability condition or not. For this reason, we consider two types of covariance matrix $ \bSigma= 0.1\times\{0.5^{|i-j|} \}_{i,j=1,\dots,p}$ and $\bSigma= \{0.5^{|i-j|} \}_{i,j=1,\dots,p}$. This setting is similar to the setting in \cite{cai2023statistical}. The former will ensure the bounded individual probability condition holds, while the latter will break this condition since the resulting probability can be pretty close to $0$.

 The performance is based on four criteria: (1) Absolute bias: the absolute distance between estimator and ground truth. (2) Coverage probability of 95\% confidence interval (3) Length of 95\% confidence interval and (4) the computation times. There are three different categories  among $\bbeta^{*}_{i}, i \in [p]$, that is $0, 1$ and $-1$. We choose three elements from each category and calculate the corresponding debiased estimator respectively. The averaged performances for each category are recorded. The experiment is conducted with 500 replications and the results are summarized in Tables \ref{n200p500sigma0.1} and \ref{n200p500sigma1}. 

Table \ref{n200p500sigma0.1} presents the results when the covariance matrix  $ \bSigma= 0.1\times\{0.5^{|i-j|} \}_{i,j=1,\dots,p}$. 
Notably,  all four methods are able to achieve the desired coverage probability. Our ADL estimator takes the minimum time and can still achieve the comparable length of the confidence interval. In Table \ref{n200p500sigma1}, when the bounded individual probability condition is violated, both deLasso and its online variant ODL fail to achieve $95\%$ coverage probability. This phenomenon has also been observed in \cite{cai2023statistical}. In contrast, LSW and ADL are still able to achieve $95\%$ coverage probability. 
This result further supports our finding that similar to LSW, the ADL does not depend on the bounded individual probability condition. Compared with LSW, the ADL estimator is around 60 to 70 times faster with similar statistical efficiency. In summary, the ADL enjoys a small time complexity and still yields high statistical accuracy.

In the second experiment, we aim to demonstrate the low space complexity of the ADL.   We increase the value of $p$ to $20000$ and choose $n=1000,  s_0 = 20 $ with $\bSigma= \{0.5^{|i-j|} \}_{i,j=1,\dots,p}$. 
Under this circumstance, the computation of deLasso, LSW, or ODL is out of memory and only ADL is computable. Similar to the first setting, we construct the confidence intervals for three randomly selected parameters from each category of $\bbeta^*$. The averaged result in each category is summarized in Figure \ref{fig:p20000}. In Figure \ref{fig:p20000}, the lines and shadow stand for the change of debiased estimator and the length of 95\% confidence interval along with the increase of data size respectively.  We observe that the estimator is close to the corresponding true value in every category, which shows the accuracy of our method. In addition, the length of the confidence interval decreases as the sample size increases, and the coverage probability is close to 95\% in each case, with detailed results provided in the supplementary material. 

 The second experiment indicates that the ADL has low space complexity while maintaining the desired statistical accuracy.

\begin{table}[htb!]
	\caption{ Simulation results  over 500 replications with $n=200$, $p=500$, $s_0=6$, $ \bSigma= 0.1\times\{0.5^{|i-j|} \}_{i,j=1,\dots,p}$.}
	\label{n200p500sigma0.1}
	\centering
	\begin{tabular}{l r| c|c|ccc|ccc}
		\hline\hline
		&\multicolumn{1}{c}{$\bbeta_{k}^*$}&\multicolumn{1}{c}{deLasso}&\multicolumn{1}{c}{LSW} &\multicolumn{3}{c}{ODL}&\multicolumn{3}{c}{ADL} \\

		\hline
		Sample size $n$ &   &200 &200 &80
		&140 &200 &80 &140&200\\
		
		\hline
		\multirow{3}{*}{\shortstack[l]{Absolute\\bias}}   &0 
		& 0.409 &0.374& 0.032& 0.029& 0.026& 1.360& 0.609& 0.486\\
		&1  
		& 0.397&0.485& 0.228& 0.180& 0.165& 1.326& 0.638& 0.497\\
		&-1  
		& 0.414&0.511 & 0.246& 0.206& 0.189& 1.308& 0.611& 0.484\\
		
		\hline
		\multirow{3}{*}{\shortstack[l]{Coverage\\probability}}   &0 
		& 0.952&0.972& 0.980& 0.971& 0.966&0.945& 0.974& 0.960\\
		&1  
		& 0.962&0.936& 0.978& 0.962& 0.962& 0.969& 0.969& 0.958\\
		&-1  
		& 0.955&0.930& 0.971& 0.958& 0.957& 0.961& 0.967& 0.965\\

		\hline
		\multirow{3}{*}{\shortstack[l]{Confidence \\ interval length}}   &0 
		& 2.032&2.200& 3.076& 2.333& 1.970& 6.503& 3.326& 2.527\\
		&1  
		& 2.034&2.197& 3.074& 2.337& 1.972& 6.461& 3.334& 2.539\\
		&-1  
		& 2.032&2.202& 3.070& 2.333& 1.969& 6.470& 3.325& 2.528\\
		
		\hline
		Time $(s)$& \multicolumn{1}{c}{}& \multicolumn{1}{c}{7.954} &\multicolumn{1}{c}{21.017} &\multicolumn{3}{c}{1.228} &\multicolumn{3}{c}{0.310} \\
		\hline\hline
	\end{tabular}
\end{table}

\begin{table}[htb!]
	\caption{Simulation results averaged over 500 replications with $n=200$, $p=500$, $s_0=6$, $ \bSigma= \{0.5^{|i-j|} \}_{i,j=1,\dots,p}$.}
	\label{n200p500sigma1}
	\centering
	\begin{tabular}{l r| c|c|ccc|ccc}
		\hline\hline
		&\multicolumn{1}{c}{$\bbeta_{k}^*$}&\multicolumn{1}{c}{deLasso}&\multicolumn{1}{c}{LSW} &\multicolumn{3}{c}{ODL}&\multicolumn{3}{c}{ADL} \\
		\hline
		Sample size $n$ &   &200 &200 &80
		&140 &200 &80 &140&200\\
		
		\hline
		\multirow{3}{*}{\shortstack[l]{Absolute\\bias}}   &0 
		& 0.115& 0.117& 0.013& 0.011& 0.010& 0.830& 0.364& 0.298\\
		&1  
		& 0.337& 0.260& 0.583& 0.515& 0.486& 0.811& 0.399& 0.311\\
		&-1  
		& 0.338& 0.256& 0.574& 0.511& 0.481& 0.774& 0.395& 0.310\\
		
		\hline
		\multirow{3}{*}{\shortstack[l]{Coverage\\probability}}   &0 
		& 0.982& 0.995& 0.999& 0.999& 0.997& 0.958& 0.973& 0.960\\
		&1  
		& 0.578& 0.928& 0.948& 0.793& 0.672& 0.952& 0.945&0.940\\
		&-1  
		&0.573 & 0.940& 0.947& 0.802& 0.666& 0.962& 0.951& 0.944\\

		\hline
		\multirow{3}{*}{\shortstack[l]{Confidence \\ interval length}}   &0 
		& 0.749& 1.206& 1.836& 1.353& 1.130& 3.868& 1.953& 1.492\\
		&1  
		& 0.763& 1.243& 1.827& 1.351& 1.128& 3.933& 1.978& 1.511\\
		&-1  
		& 0.763& 1.245& 1.833& 1.347& 1.129& 3.890& 1.967& 1.502\\
		
		\hline
		Time $(s)$& \multicolumn{1}{c}{}& \multicolumn{1}{c}{8.127} & \multicolumn{1}{c}{19.744}&\multicolumn{3}{c}{1.189} &\multicolumn{3}{c}{0.293} \\
		\hline\hline
	\end{tabular}
\end{table}

\begin{figure}[h!]
    \centering
    \includegraphics[width=\textwidth]{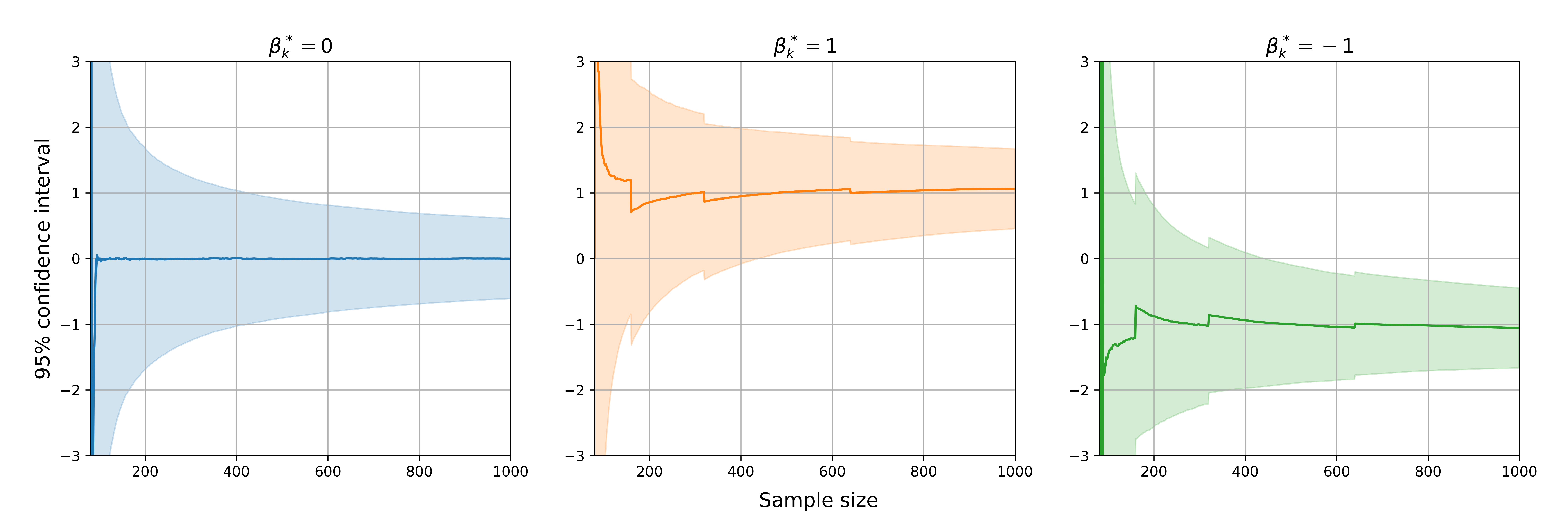}
       \caption{Simulation results averaged over 500 replications (each takes around 42.452s) with $n=1000$, $p=20000$, $s_0=20$, $ \bSigma= \{0.5^{|i-j|} \}_{i,j=1,\dots,p}$.} \label{fig:p20000}
\end{figure}
\spacingset{1.9}

\section{Real data example: spam email classification}\label{sec:real}

With the advancement of digital communication, email has become an indispensable tool for both personal and professional correspondence. However, this widespread adoption has also attracted malicious actors who exploit email systems to distribute spam, phishing attempts, and fraudulent content. Identifying spam emails and understanding the features that characterize them are crucial for protecting users from potential security threats, financial fraud, and privacy breaches.

For this purpose, we analyze a dataset on spam email classification\footnote{\url{https://www.kaggle.com/datasets/purusinghvi/email-spam-classification-dataset}}. 
This dataset consists of 83,446 email messages with binary labels for spam or non-spam.  We construct features by counting the frequencies of unigrams and bigrams from the email content and then applying a log transformation.  In addition, we partition the data into training and test sets to evaluate our algorithm's prediction accuracy.

Specifically, we split the data into 80\% for training and 20\% for testing.
As a result, the feature dimension $p\approx 3.6$ million, and the training data size $n\approx 66,000$. We then apply our method to the training set and evaluate the prediction accuracy on the test set using the estimator derived from the training step. It is important to note that this dataset is ultra-high dimensional and the existing offline methods are computationally prohibitive, so we only show results from our proposed ADL method.

We select the terms of our interests: ``investment", ``schedule", and ``per cent" to conduct statistical inference, with their confidence bands shown in the upper left, upper right, and bottom left of Figure \ref{fig:spam}. For each term, the corresponding confidence interval gets narrower as more data are processed. Notably, the term ``investment" consistently shows a positive coefficient, with confidence intervals above zero after processing 950 data points. In contrast, the term ``schedule" presents a significantly negative relationship, with its confidence intervals falling below zero after 1100 data points. These findings align with our intuition: emails containing ``investment" are more likely to be financial scams, while ``schedule" typically appears in trusted business communications, like scheduling meetings.  Unlike ``investment" and ``schedule," the term ``per cent" shows a neutral influence, with confidence bands covering zero. This phenomenon suggests that ``per cent'' is a common term in both spam and non-spam emails. Thus, it has a limited ability to distinguish spam from non-spam emails. The bottom-right panel of Figure \ref{fig:spam} shows the prediction error rate evaluated on the test set as training sample size increases. As the ADL estimator is updated alongside the increasing training sample, the prediction error rate steadily decreases from approximately 0.47 to 0.15. This result demonstrates the effectiveness of our method from the perspective of prediction accuracy.

\spacingset{1.25}

\begin{figure}[htb]
    \centering
    \includegraphics[trim=1.5cm 1cm 2cm 0cm, width=0.85\textwidth]{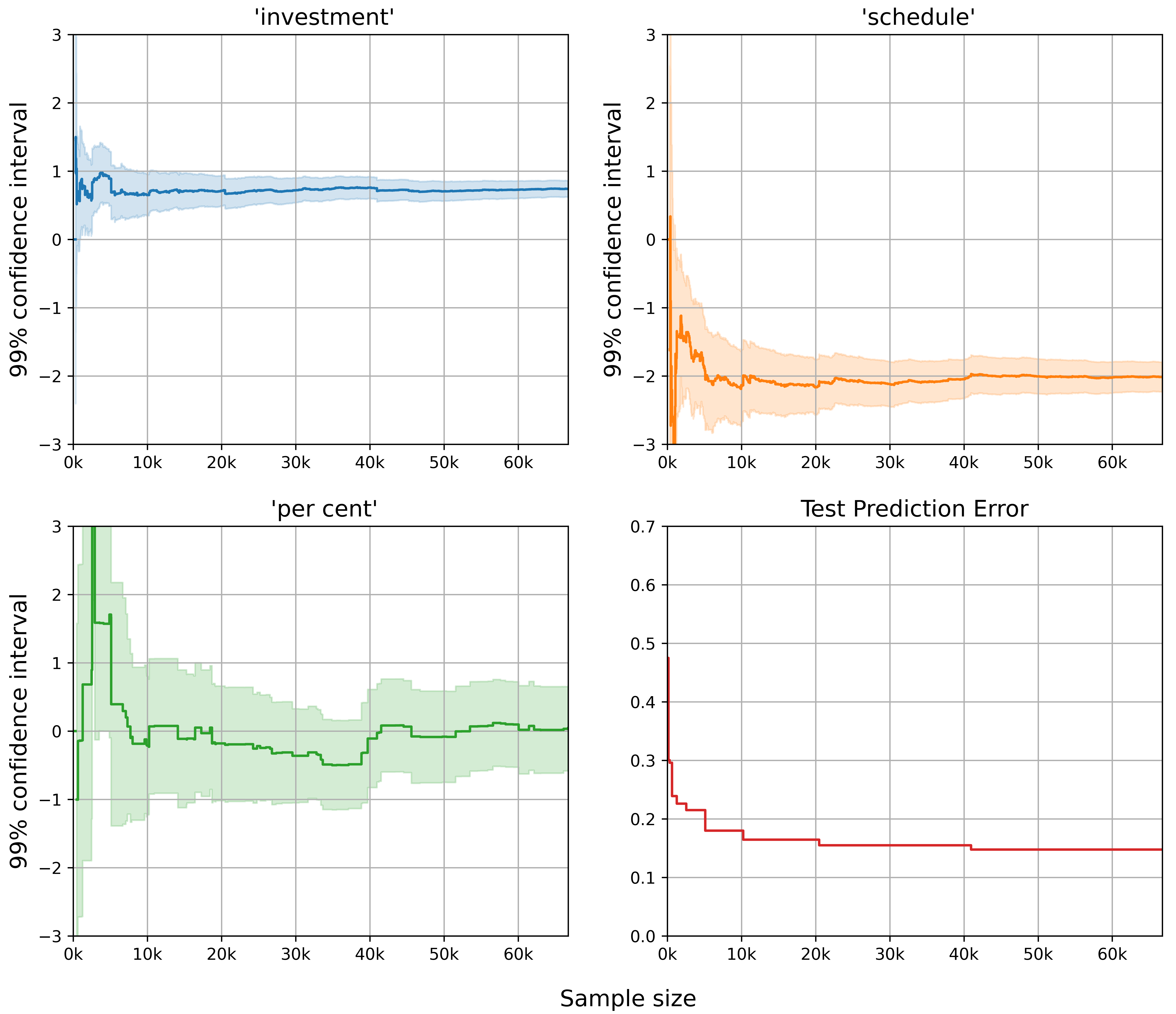}
 
    \caption{99\% confidence interval estimate for ``investment", ``schedule", and ``per cent" (upper left, upper right, and bottom left), where lines and shaded areas represent the traces of the ADL estimator and  99\% confidence interval respectively. The prediction error evaluated on the test set, as training sample size increases, is shown in the bottom right.}   
    \label{fig:spam}
\end{figure}

\spacingset{1.9}

\section{Conclusion}\label{sec:conc}
High-dimensional inference has been well investigated in the last decade. However, it still lacks an efficient method to conduct the inference in a real online manner. In this work, we contribute to the literature by proposing the ADL to deal with the high-dimensional generalized linear models. While maintaining comparable accuracy to the offline method, the ADL requires much less time and space complexities. 

Several directions deserve further investigation. First, the ADL is designed for an objective function with a lasso penalty. It is unclear how it would perform with other penalty functions, such as SCAD \citep{fan2001variable} and MCP \citep{zhang2010nearly}. 
 Our initial numerical explorations, presented in the supplementary material, indicate that the proposed adaptive debiasing procedure can still deliver satisfactory performance when using SCAD or MCP.  However, the theoretical justifications of the adaptive debiased estimator with SCAD or MCP are rather challenging, especially when non-convexity of the penalty is introduced into the objective function. Second, our work focuses on the parametric models. It is interesting to investigate whether our approach can be extended to the model with a non-parametric component \citep{fang2023online,quan2024optimal} to conduct semi-parametric inference.
Last but not least, this work assumes that sequentially collected data points are independent and identically distributed (i.i.d). How to deal with  non-i.i.d. design, such as the covariate shift, under an online setting is still an open question.

\begin{description}
	\item[Supplementary material:] We include detailed proofs of Theorems \ref{bbeta_oracle}--\ref{asy_normal}. In addition, we conduct extensive simulation studies, including different covariance matrices,  ablation studies, and power analysis. Furthermore,  we provide practical guidance for selecting hyper-parameters in our algorithm, along with a sensitivity analysis.
	\vspace{-0.3cm}
	\item[Code:] We include the Python code in a GitHub repository to reproduce the results in the numerical studies, \url{https://github.com/ChattelionLuo/ADL}.
\end{description}

\spacingset{1.9}

\bibliographystyle{apalike}
\bibliography{paper-ref}

\end{document}